%% file: main.tex
\documentclass[runningheads]{llncs}

\usepackage{eccv}

\usepackage{eccvabbrv}

\usepackage{graphicx}
\usepackage{booktabs}
\usepackage{wrapfig}
\usepackage{xfrac}

\usepackage[accsupp]{axessibility}  

\usepackage[pagebackref,breaklinks,colorlinks,citecolor=eccvblue]{hyperref}

\usepackage{orcidlink}

\begin{document}

\title{YaART: Yet Another ART Rendering Technology
} 

\titlerunning{YaART: Yet Another Art Rendering Technology}

\author{
Sergey Kastryulin\inst{1,2} \and
Artem Konev\inst{1}$^*$ \and
Alexander Shishenya\inst{1}$^*$ \and \\
Eugene Lyapustin\inst{1}$^*$ \and 
Artem Khurshudov\inst{1}$^*$ \and
Alexander Tselousov\inst{1} \and \\
Nikita Vinokurov\inst{1} \and 
Denis Kuznedelev\inst{1,2} \and
Alexander Markovich\inst{1} \and \\
Grigoriy Livshits\inst{1} \and
Alexey Kirillov\inst{1,3} \and
Anastasiia Tabisheva\inst{1} \and \\
Liubov Chubarova\inst{1,4} \and
Marina Kaminskaia\inst{1} \and
Alexander Ustyuzhanin\inst{1} \and \\
Artemii Shvetsov\inst{1} \and
Daniil Shlenskii\inst{1,2} \and
Valerii Startsev\inst{1} \and \\
Dmitrii Kornilov\inst{1} \and
Mikhail Romanov\inst{1} \and \\
Artem Babenko\inst{1}$^\dagger$ \and
Sergei Ovcharenko\inst{1}$^\dagger$ \and
Valentin Khrulkov\inst{1}$^\dagger$
}

\authorrunning{S.~Kastryulin \textit{et al.}}

\institute{Yandex \and
Skolkovo Institute of Science and Technology \and
Moscow State University \and
Higher School of Economics
}
\maketitle

\def\thefootnote{*}\footnotetext{Equal contribution}
\def\thefootnote{$\dagger$}\footnotetext{Equal senior authors}
\input{sections/00_abstract}
\input{sections/01_intro}
\input{sections/02_related}

\input{sections/03_approach}
\input{sections/04_experiments}
\input{sections/10_conclusion}

\clearpage  

%
%

\bibliographystyle{splncs04}
\bibliography{main}

\appendix 
\input{supmat/01_intro}
\input{supmat/05_pretrain_ft_qvq}
\input{supmat/06_limitations}
\input{supmat/07_more_examples}

\end{document}

%% file: sections/00_abstract.tex
\begin{abstract}
In the rapidly progressing field of generative models, the development of efficient and high-fidelity text-to-image diffusion systems represents a significant frontier. This study introduces \textbf{YaART}, a novel production-grade text-to-image cascaded diffusion model aligned to human preferences using Reinforcement Learning from Human Feedback (RLHF). During the development of \textbf{YaART}, we especially focus on the choices of the model and training dataset sizes, the aspects that were not systematically investigated for text-to-image cascaded diffusion models before. In particular, we comprehensively analyze how these choices affect both the efficiency of the training process and the quality of the generated images, which are highly important in practice. Furthermore, we demonstrate that models trained on smaller datasets of higher-quality images can successfully compete with those trained on larger datasets, establishing a more efficient scenario of diffusion models training. From the quality perspective, \textbf{YaART} is consistently preferred by users over many existing state-of-the-art models.

\keywords{Diffusion models \and Scaling \and Efficiency}
\end{abstract}

%% file: sections/01_intro.tex
\section{Introduction}
\label{sec:intro}

\begin{figure}[!ht]
    \centering
    \makebox[\textwidth][c]{\includegraphics[width=1.5\linewidth]{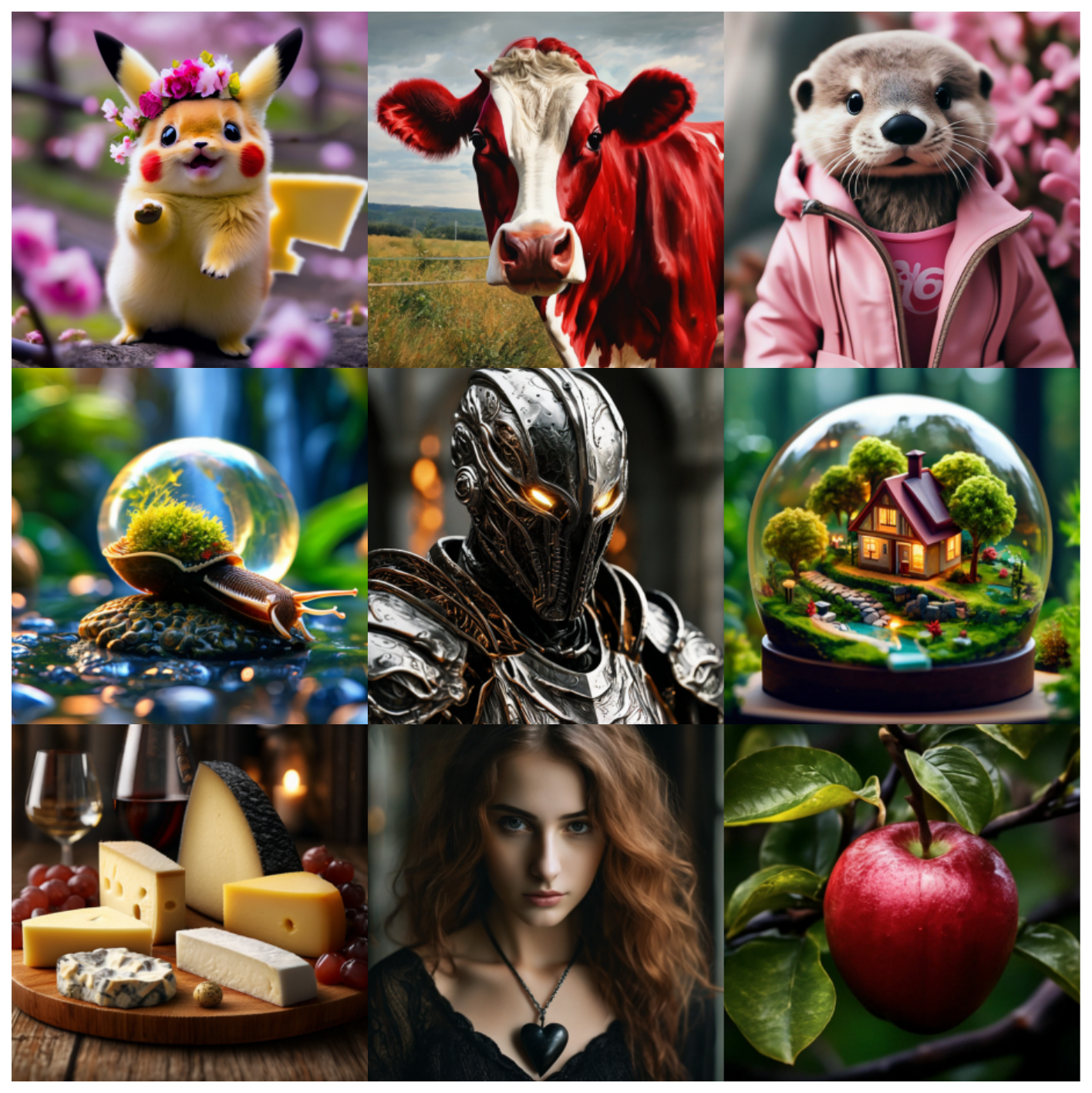}}
    \caption{RL-aligned YaART generates visually pleasing and highly consistent images.}
    \label{fig:collage}
    \vspace{-16pt}
\end{figure}

Large-scale diffusion models have recently achieved unprecedented success in text-conditional image generation, which aims to produce high-quality images closely aligned with user-specified textual prompts. 
The state-of-the-art models \cite{chen2023pixart, dai2023emu, rombach2022_LDM, arkhipkin2023kandinsky3, betker2023_dalle3, pernias2023wurstchen, podell2023sdxl} can generate complex, photorealistic images, which paves the way to their ubiquitous usage in applications, including web design, graphics editors, e-commerce and others. 

Outstanding image generation ability typically comes at a price of scale, i.e., state-of-the-art systems require large models, large training datasets, and vast amounts of GPU resources. 
At the moment, it is not completely clear what the trade-offs are between these three aspects of scale and the generation quality. 
Moreover, it is unclear if one should prefer large datasets of random web images over smaller datasets containing only well-filtered, high-quality ones.

In this paper, we introduce YaART --- a cascaded text-to-image diffusion model enhanced with human preference tuning. 
YaART is specifically designed to optimize data and computational resource usage. We thoroughly explore the impact of model and dataset sizes on generation performance and training costs.
Furthermore, we investigate the quality-vs-quantity trade-off for the training dataset construction and reveal that training on smaller datasets of high-fidelity samples provides performance on par with models trained on larger datasets.
Additionally, we show that the model size can be successfully traded for training time, i.e., smaller models can perform on par with the larger ones if their training time budget is long enough. 
Regarding quality, YaART is consistently preferred in human evaluations over well-known benchmarks, such as SDXL v1.0, MidJourney v5, Kandinsky v3, and OpenJourney.  

To sum up, our main contributions are:
\begin{itemize}

    \item We introduce YaART, an RLHF-tuned cascaded diffusion model that outperforms the established models regarding side-by-side human evaluation of image realism and textual alignment. 
    
    \item We systematically study the influence of interactions of model and dataset sizes on the training efficiency and the final model quality. To the best of our knowledge, this is the first work to extensively investigate the scalability of diffusion models in convolution-based cascaded schemes using a reliable human evaluation protocol.
    
    \item We investigate the practically important trade-off between data quality and quantity for pre-training diffusion models. We reveal that competitive results can be obtained by training on a small amount of high-quality data.

    \item To the best of our knowledge, we are the first to thoroughly describe RLHF-tuning of the production-grade diffusion model. We show that RLHF tuning is an important ingredient of advanced performance; in particular, it substantially improves image aesthetics and reduces visual defects.
\end{itemize}

%% file: sections/02_related.tex
\section{Related work}
\label{sec:related}
Despite the recent progress on large-scale diffusion models, achieving high-fidelity synthesis of high-resolution images remains a formidable challenge. 
Two primary approaches address this issue. The first, employed in models like DALLE-2 \cite{ramesh2022hierarchical} and Imagen \cite{saharia2022_Imagen}, termed `cascaded', involves a sequential pipeline of multiple diffusion models that progressively generate higher-resolution images. 
The second approach, known as the Latent Diffusion Model \cite{rombach2022_LDM}, leverages score matching on the latent space of images generated by a pre-trained VAE.
This method has yielded impressive results and has been adopted in various subsequent works such as Stable Diffusion XL \cite{podell2023sdxl}, DALL-E 3 \cite{betker2023_dalle3}, PixArt-$\alpha$ \cite{chen2023pixart}, EMU \cite{dai2023emu}. 

Traditionally, the denoiser network in diffusion models is implemented as a variant of the convolutional U-Net model \cite{ronneberger2015u} with additional attention blocks. 
However, inspired by the success of the transformer architecture \cite{vaswani2017attention} in language modeling, recent research \cite{peebles2023_DiT,bao2023all,chen2023pixart,zheng2023fast} has replaced the convolutional backbone with a transformer backbone, enabling straightforward parameter expansion.

In our paper, we adhere primarily to the original approach proposed in DALL-E 2 and Imagen. 
Specifically, we employ a cascaded scheme of three diffusion models and a convolutional backbone for denoisers.

\subsubsection{Alignment of Text-to-Image Models on Human Preferences.}

Fine-tuning diffusion models on a clean filtered dataset has been shown to drastically improve the quality of generated images; however, improving quality even beyond the level of the available datasets requires other training methods. 
Approaches like \cite{kimin2023} and \cite{kevin2023_DRaFT} suggest generating a dataset with the current diffusion model, ranking them with scoring models, rejecting samples with low scores, and fine-tuning diffusion model on this dataset with MSE loss or reward-weighted MSE loss. In \cite{ying2023_DPOK}, \cite{kevin2023_DDPO}, and \cite{seung2024_Parrot}, authors adapt REINFORCE and PPO for tuning diffusion models. Papers \cite{kai2023_D3PO} and \cite{bram2023_Diffusion_DPO} exploit DPO for improving quality of generated images, which allows to avoid training explicit reward functions. Some methods like \cite{kevin2023_DRaFT} rely on differentiable scoring functions while tuning diffusion models.

Diffusion models require multiple steps of inference for generating a single image; thus, methods like \cite{kevin2023_DRaFT} that utilize differentiable rewards and backpropagate through these rewards and all the iterations of image generation face a couple of significant problems: vanishing gradients and large GPU memory requirements for each sample. RL-based approaches like \cite{ying2023_DPOK}, \cite{kevin2023_DDPO}, and \cite{seung2024_Parrot} require agents to generate trajectories online during training, which induces additional computational cost. Besides, these methods use on-policy algorithms, which do not allow them to use images from any sources but the current model itself. DPO-based approaches like \cite{kai2023_D3PO} perform on par with D3PO but have the advantage that they can exploit samples from any other sources.


\subsubsection{Scaling of Diffusion Models.}

The remarkable progress of text-to-image generative models is largely driven by improvements in scale, where bigger models are trained on larger datasets for longer schedules.
However, the amount of research aimed to understand the scalability of text-to-image models is still limited. 
Saharia \textit{et al}. \cite{saharia2022_Imagen} investigate the influence of U-Net model size on image quality and conclude that increasing the model size has little effect, whereas scaling the text encoder has a greater impact. 
Caballero \textit{et al.} \cite{caballero2022broken} build a power scaling law from experiments on ImageNet 64x64 \cite{deng2009_imagenet}, and LSUN Bedrooms 256x256 \cite{yu15_lsun} datasets for single-step diffusion models, confirming the potential benefit of size and compute scaling.
Diffusion Transformer (DiT) \cite{peebles2023_DiT} study the effectiveness of usage of transformer models in the Latent Diffusion \cite{rombach2022_LDM} framework and finds that larger DiT models are more computationally efficient and that scaling the DiT improves generation quality in all stages of training. 

All mentioned studies mainly rely on FID \cite{heusel2017_FID} for image quality estimation, which has been shown to correlate poorly with human judgments \cite{borji2019_pros_and_cons_of_gan_metrics, sajjadi2018_precision_recall_metric}.
Moreover, the questions of data quality and quality remain unexplored.
In this work, we study the scalability of the convolutional diffusion model in the cascade framework based on large-scale study with human annotators.
We investigate the influence of model and dataset sizes on end quality, specifically considering the data quality-quantity trade-off.

%% file: sections/03_approach.tex
\section{Approach}
\label{sec:approach}

Here, we describe our approach to train large-scale diffusion models. 

\subsubsection{High-resolution image synthesis.} 
We initially opted for the cascaded pixel-based diffusion approach because of several considerations. 
Firstly, by leveraging a pre-trained VAE to extract latent codes for LDM, the quality of the generated images is inherently constrained by the VAE's performance, which often has limited capacity. Although the VAE could be fine-tuned later, such adjustments might not substantially enhance its overall capability due to inherent capacity limitations and difficulties with training VAE-based models.

In contrast, cascaded diffusion models allow for important practical advantages. In cascaded models, the ``decoder'' function is a stack of two super-resolution diffusion models that operate independently of the primary pixel-space generative model. This autonomy allows for the natural interchangeability of the super-resolution blocks. Furthermore, the diffusion decoder's increased capacity and enhanced generative capability significantly contribute to improving the fidelity of the resultant images.

\subsubsection{Large high-quality datasets for model pre-training.} 
One of the most difficult challenges in training a large-scale text-to-image model is curating a high-quality dataset of text-image pairs. 
We have continually refined the process of assembling such a dataset, ultimately converging on a multi-stage procedure. 
This approach encompasses separate pre-filtration stages of image and text filtration and the final stage of selecting image-text pairs according to their overall fidelity with a focus on relevance.

\subsubsection{Boosting relevance and quality via supervised fine-tuning.}
Pre-trained text-to-image models often encounter difficulties in image generation. 
Exact prompt following, in combination with high visual attractiveness, is often a challenge. 
Our strategy involves supervised fine-tuning using a manually selected subset of data with exceptional aesthetic quality and detailed descriptions. 
This process notably enhances prompt following and relevance, focusing on the latter.

\subsubsection{Aesthetic refinement through Reinforcement Learning.}
In the final stage, we further elevate image aesthetics and quality through Reinforcement Learning (RL). 
We directly use the signal provided by human annotators to improve the aesthetic properties of the produced images.
\\[12pt]
To sum up, our approach has three main stages. 
First, we pre-train a diffusion model on a large-scale dataset, followed by the fine-tuning stage. This model is further polished via RL-tuning, which incorporates human feedback. 
Next, we discuss each component in detail.


\subsection{Cascaded Diffusion}
\label{sec:model}
As discussed above, we base our approach on the framework of cascaded diffusion models. 
Specifically, we follow Imagen \cite{saharia2022_Imagen} to design the cascaded diffusion model. 
The initial diffusion model in the cascade, which produces $64 \times 64$ images (GEN64), and the first super-resolution model, which performs up-sampling $64 \to 256$ (SR256), follow the architecture of U-Net in \cite{dhariwal2021diffusion,rombach2022_LDM,saharia2022_Imagen,ramesh2022hierarchical}.
The GEN64 is conditioned on text via the cross-attention mechanism as in \cite{rombach2022_LDM}. 
The SR256 model is purely convolutional with a single self-attention in the middle block. 
Additionally, both models are conditioned on text via modulation of residual blocks by linear projections of the mean-pooled text embeddings (as described in \cite{ramesh2022hierarchical}). 
For the final super-resolution stage $256 \to 1024$ (SR1024), we utilize the text-unconditional Efficient U-Net backbone with the same configuration as the third stage in \cite{saharia2022_Imagen}.
Similarly to Imagen, we utilize large pre-trained text encoders for conditioning. 
Specifically, we use a CLIP \cite{radford2021learning}-like proprietary model of size $\sim 1.3B$ which follows BERT \cite{devlin2018bert}-xlarge architecture. 
It was pre-trained on a mixture of public and proprietary datasets of text-image pairs in a setup similar to OpenAI CLIP models and demonstrated competitive performance on public benchmarks. 
Our preliminary smaller-scale experiments verified that this text encoder provides better quality of generated images according to human raters than the other choices such as CLIP/OpenCLIP\cite{cherti2023reproducible}/FlanT5-XXL \cite{chung2022scaling} encoders. 
In all our experiments, we use the standard linear noise scheduler\cite{dhariwal2021diffusion} with continuous time; we follow the implementation in \cite{kingma2021variational}. 
The timestep $t$ is provided to the model via its respective $LogSNR(t)$, computed by the scheduler.

\subsection{Data Selection Strategy}
\label{sec:data-selection}

Our proprietary pool of images is comprised of 110B image-text pairs. 
For each pair we have a pre-calculated set of scores produced by multiple learned predictors.
These predictors are fully connected classifiers, trained on top of image and text features, extracted by our proprietary visual and textual foundation models. 
For example, we have various classifiers for image aesthetics and quality, watermark detectors, NSFW detectors, a binary classifier to check the presence of text within an image, and more.

\subsubsection{Selecting the best images.}

We formulate several stages of filtering for the initial pool of images, starting from defining the \textbf{Image Score} as a linear combination of the image-based predictors with weights tuned on the SAC \cite{pressmancrowson2022_SAC} dataset.
By visual inspection in the preliminary experiments, we confirm that the visual appeal of the images is correlated with the \textbf{Image Score} predictions.
We decided to keep only the top \sfrac{1}{3} of the images because most of the images below this threshold had low visual attractiveness.
Additionally, we remove NSFW images utilizing our proprietary classifiers and take ones with a size in the range of [512, 10240] pixels and an aspect ratio in the range of [0.5, 2].
These values were chosen to filter out images with unusual sizes that may increase computational and engineering burden on our image processing pipeline or have undesirable content.
Then, we filter out images that do not pass thresholds for the image quality classifiers, specifically those that contain high levels of noise, blurriness, watermarks, or a checkered background and were subjected to a high degree of compression.
Finally, we manually select thresholds for aesthetics classifiers, trained on AVA \cite{murray2012ava} and TAD66k \cite{he2022_TAD} datasets.
We also explicitly control background monotonicity by splitting the dataset into two parts and allowing only 10\% of the samples during training to have a monotonic background.
The motivation is to emphasize images rich with details during training while retaining the ability to generate mono backgrounds if required.


Our SR1024 model does not have a condition on the texts. Hence, we obtain the final version of the 180M Main SR dataset by further tightening the filtering conditions at this stage, especially those related to image quality.

\subsubsection{Text filtration.}

First, we focus on English texts only by recognizing text language with our proprietary language detector. 
Next, we manually annotate a random subset of 4.8K texts so that each raw input text receives its cleaned-up version or an empty mark if deemed unsuitable for training. Finally, we compile the training dataset from the refined versions of the texts, filtering out empty text entries. We then fine-tune our proprietary 180M Language Model on that dataset and use its prediction as a factor of text quality.


\subsubsection{Combining All Together.}

Previously, we discussed the image filtering process and computation of various image and text factors.
At this point, we have a pool of 2.3B of relatively poor-quality data, which we aim to filter out further.
For that, we manually label 66K image-text pairs for their visual attractiveness, textual descriptiveness, and relevance on a scale from 1 to 3.
Unlike the more frequently used Likert scale, the simplified scale balances the descriptiveness of the score with the noise in the votes of the assessors. 
After that, we train a CatBoost \cite{prokhorenkova2018catboost} model on a set of 56 factors, among which there are 6 variations of CLIP scores, 38 text-only, and 12 image-only factors, to obtain Sample Fidelity Classifier (SFC).


All image-text pairs from the previous filtration stage are sorted according to the SFC model prediction, and the top pairs are selected as the final training pool.
We select a threshold so the final dataset has 300M images with a non-monotonic background.
To obtain the final 330M Main Gen dataset for pre-training, we also add a random sample of 30M pairs containing images with monotonic backgrounds.

\subsection{Model Training}
\label{sec:model_training}

\subsubsection{Pre-training.}
The model is initially pre-trained on a dataset collected as described above. The GEN64 model with 2.3B parameters is trained on $160$ A100-80GB GPUs with the total batch size being $4800$ for $1.1 \cdot 10^6$ training iterations.
The model is trained with automated mixed-precision using ZeRO \cite{rasley2020deepspeed}. 
We use Adam optimizer with learning rate being $10^{-4}$ and $\beta_1=0.9, \beta_2=0.98$; we observed that the smaller values of $\beta_2$ enhance training stability in half precision. 
The SR256 model with 700M parameters is trained similarly on $80$ A100-80GB GPUs with total batch size being $960$ and reduced learning rate $5 \times 10^{-5}$; we train it for $1.5 \times 10^6$ iterations. 
The SR1024 EfficientUNet \cite{saharia2022_Imagen}, with 700M parameters, is trained on 256 $\times$ 256 crops under a similar setup, albeit with a smaller batch size of $512$.
We summarize the main information about training in Table \ref{tab:datasets}.
Detailed model configurations are provided in the supplementary material.

\subsubsection{Fine-tuning.}
\label{sec:fine-tuning}
To fine-tune the GEN64 and SR256 models, we manually collected a dataset of 50K samples consisting of high-quality images and relevant descriptive captions. 
At this stage, we value relevance more than image attractiveness, which we align later via RL.
We also found that fine-tuning the SR1024 model on full-resolution images from the 180M Main SR dataset leads to a noticeable improvement in the fine-grained details and sharpness of the images.

\input{tables/datasets}

\subsubsection{RL Alignment.}

For RL alignment, we exploit the PPO \cite{schulman2017_PPO} approach with \( \epsilon=0.5 \) for loss computation and follow DDPO with importance sampling \cite{kevin2023_DDPO}.
We also use a value function that is trained with MSE loss. 
However, we do not use KL-divergence and classifier-free guidance because, in our experiments, they worsen convergence speed, reward growth, and slow down the generation of trajectories.

For text-to-image diffusion models, the full probability density \( p_{\theta} \left( x_{t-1} | x_t, c \right) \) is a product of multiple Gaussian probability density functions for each pixel and channel.
It causes rapid deviation from the value of \( p_{\theta_{old}} \left( x_{t-1} | x_t, c \right) \), which leads to a significant loss of precision. 
To cope with this problem, we split each sample \( x_t \) into patches and compute the objective for each patch independently, followed by averaging over all the patches. 

We use three reward models: Relevance-focused OpenCLIP ViT-G/14 \cite{ilharco_gabriel_2021_open_clip} and predictors of image Consistency (lack of defects) and Aesthetics, trained on our human preference data in a Siamese \cite{koch2015siamese} manner with hinge loss as a fully-connected head over our in-house image foundation model.
We independently compute loss for each reward, followed by averaging them with weights.
We aim to ensure the rise of Aesthetics and Consistency rewards without significantly declining the Relevance reward. 
We experimentally found that weighting, resulting in the same rewards scale, works well in our setup.

To reduce the latency for saving and reading model updates, we train the condition encoder of our model and LoRA \cite{Edward2021_LORA} for the remaining layers. 
We also tune the frequency of the updates so they occur often enough, as it is crucial for the on-policy algorithms.
We use 100 steps of p-sample \cite{NEURIPS2020_ddpm} so the agent produces $100 \times \text{batch size}$ samples for the loss optimization.

\input{tables/sbs_baselines}

\subsection{YaART}
\label{sec:ya_art_main}

\begin{figure}[tp]
  \centering
  \includegraphics[width=.99\linewidth]{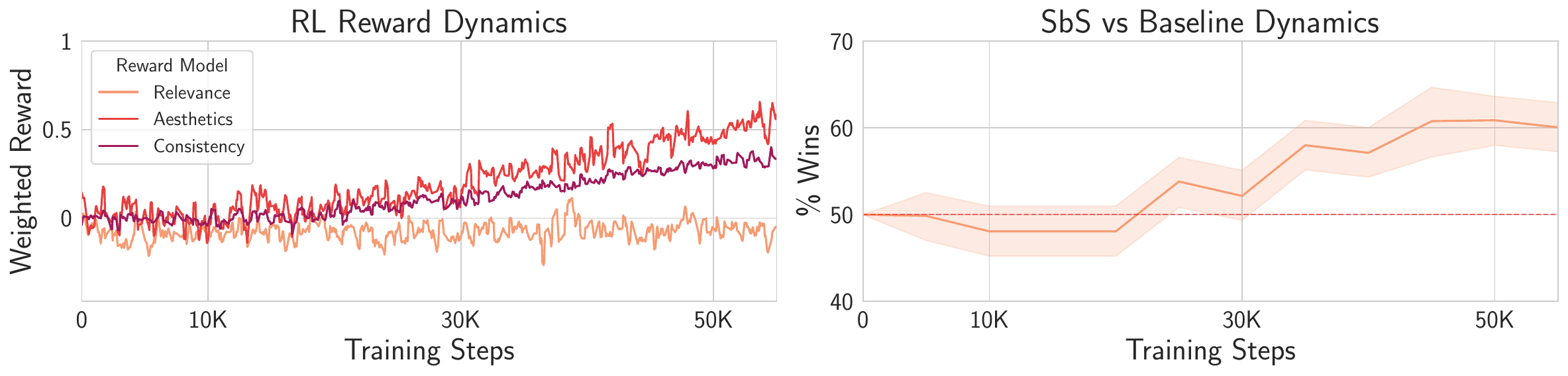} 
  \caption{An evolution of rewards (left) leads to an increase of human preference rate (right) throughout the RL alignment stage.}
  \label{fig:rl_dynamics}
\end{figure}

Following the pre-training, fine-tuning, and RL alignment pipeline, we obtain the final YaART model.
While the combination of pre-training and fine-tuning is relatively common in the modern diffusion training pipelines, we emphasize the importance of the RL alignment stage, which pushes generation quality even further. In particular, Figure \ref{fig:rl_dynamics} shows that throughout this stage, the model learns to generate more aesthetic and consistent (less defective) images while preserving image-text relevance tuned on the fine-tuning stage.

We compare YaART with MidJourney v5, Stable Diffusion XL with the refiner module, Kandinsky v3, and OpenJourney.
Table \ref{table:sbs-baselines} demonstrates that YaART performs on par with MidJourney v5 and is generally preferred over other counterparts.

%% file: tables/datasets.tex
\begin{table}[t]
\centering

\scalebox{0.7}{
\begin{tabular}{l@{\hskip .2in}c@{\hskip .2in}c@{\hskip .2in}c@{\hskip .2in}c@{\hskip .2in}c@{\hskip .2in}c@{\hskip .2in}c@{\hskip .2in}c}
\toprule
Model & Training Stage & Image Resolution & \#Samples & Training Steps & Batch Size & Learning Rate \\
\midrule
GEN64 & Pre-training & 64x64 & 330M Main Gen & 1100K & 4800 & 1 $\times 10^{-4}$ \\
SR256 & Pre-training & 256x256 & 330M Main Gen & 1500K & 960 & $5 \times 10^{-5}$ \\
SR1024 & Pre-training & 1024x1024 (crops) & 180M Main SR & 1000K & 512 & $5 \times 10^{-5}$ \\
\midrule
GEN64 & Fine-tuning & 64x64 & 50K & 40K & 240 & $1 \times 10^{-5}$ \\
SR256 & Fine-tuning & 256x256 & 50K & 40K & 96 & $1 \times 10^{-5}$ \\
SR1024 & Fine-tuning & 1024x1024 & 180M Main SR & 200K & 122 & $2 \times 10^{-6}$ \\
\midrule
GEN64 & RL-alignment & 64x64 & 300K & 50K & 192 & $1 \times 10^{-4}$ \\
\bottomrule
\end{tabular}
}
\bigskip
\caption{
A summary of training stages of the YaART model cascade.
Note that Main Gen and Main SR are large-scale datasets designed with an intent to cover a broad range of concepts while the fine-tune datasets consist of carefully selected image-text pairs, collected with a desire to boost specific quality aspects such as text-image relevance and image aesthetics.
}
\vspace{-23pt}
\label{tab:datasets}
\end{table}

%% file: tables/sbs_baselines.tex
\begin{table}[t]
\setlength\tabcolsep{2.5pt}
\begin{center}
\resizebox{.99\columnwidth}{!}{ 
\begin{tabular}{l@{\hskip .3in}c@{\hskip .2in}c@{\hskip .2in}c@{\hskip .4in}c}
\toprule

&\multicolumn{4}{c}{Quality Aspects}  \\
 \cmidrule{2-5} 
 YaART 2.3B RL & Beauty & Defects & Alignment &  Overall \\
  \cmidrule{1-5}
  vs MidJourney v5 & $0.58\pm0.03$ & $0.49\pm0.01$ & $0.52\pm0.01$ & $0.51\pm0.01$ \\
  vs SDXL & $0.75\pm0.03$ & $0.68\pm0.02$ & $0.53\pm0.02$ & $\textbf{0.77}\pm \textbf{0.01}$ \\
  vs Kandinsky v3 & $0.69\pm0.07$ & $0.69\pm0.03$ & $0.46\pm0.03$ & $\textbf{0.72}\pm\textbf{0.04}$ \\
  vs OpenJourney & $0.83\pm0.09$ & $0.75\pm0.04$ & $0.71\pm0.01$ & $\textbf{0.89}\pm\textbf{0.01}$ \\
  
\bottomrule
\end{tabular}
}
\end{center}

    
\caption{We compare YaART with state-of-the-art based on three main evaluation criteria and overall human preferences. \textbf{Bold} denotes statistically significant difference in overall models' quality.}

\label{table:sbs-baselines}
\end{table}

%% file: sections/04_experiments.tex
\section{Experiments}
\label{sec:exps}

In this section, we first describe the model evaluation approach used in our experiments. 
After that, we report the results of a systematic analysis that provides an in-depth understanding of training dynamics for different choices of data and model sizes, as well as the trade-off between data quality and quantity.
Finally, we discuss the relationship between the quality of pre-trained models and their fine-tunes.

\subsection{Evaluation Setting}
\label{sec:evaluation-setting}


\subsubsection{Prompts.}
\begin{figure}[tp]
    \centering
    \includegraphics[width=.85\linewidth]{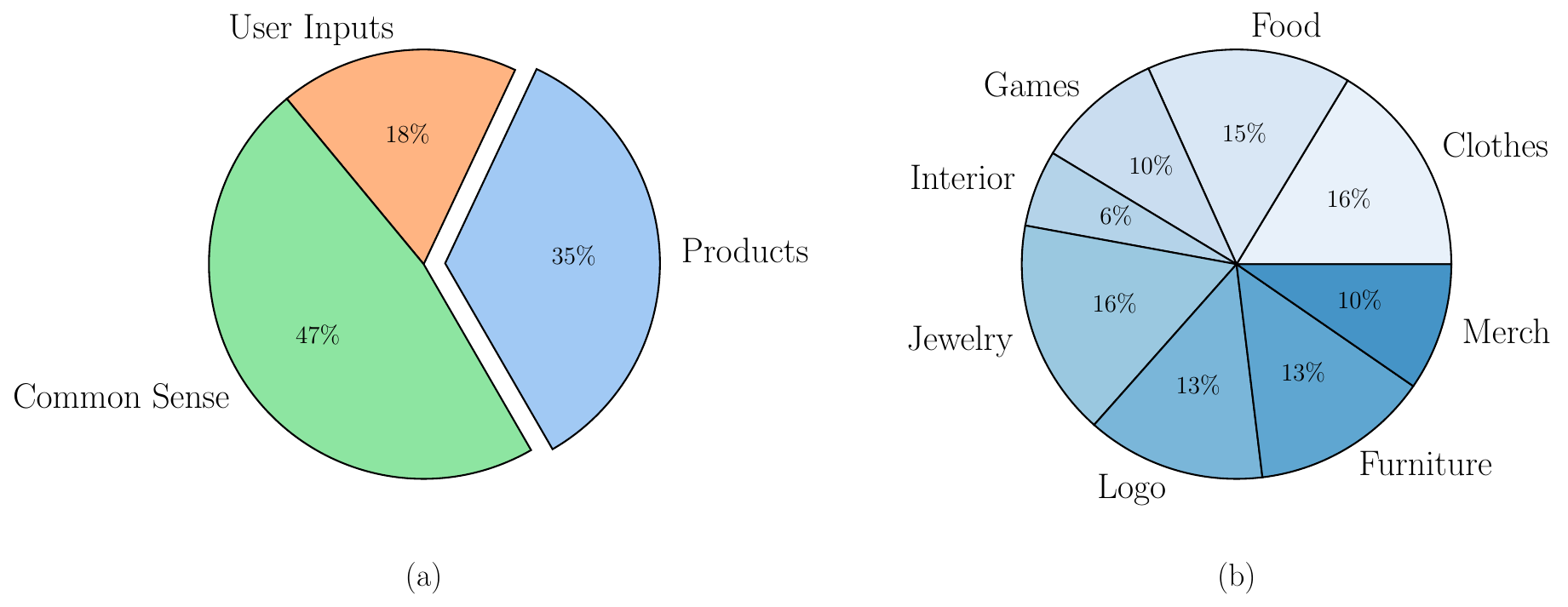}
    \caption{The content of YaBasket.
    The three major prompt categories (left) include Products, almost equally split into eight sub-categories (right).
    }
    \vspace{-13pt}
    \label{fig:yabasket}
\end{figure}

For evaluation, we use two prompt sets: our practice-driven YaBasket-300 (Figure \ref{fig:yabasket}) and DrawBench \cite{saharia2022_Imagen} -- a de-facto standard in the field of text-to-image models. 
YaBasket complements public benchmarks with several important practical use cases that have not been previously covered. Specifically, we enrich small subsets of PartiPrompts \cite{yu2022_PartiPrompts} and Winoground \cite{thrush2022_winoground}, placed into the Common Sense category, with popular requests from users of image generation models and products-driven prompts.
For completeness, we perform comparisons with established reference models in \ref{sec:ya_art_main} using both prompt sets, while all remaining sections use only the YaBasket set. We publish the prompts from YaBasket for their subsequent usage in the community\footnote{Prompts and additional information are available on the \href{https://ya.ru/ai/art/paper-yaart-v1}{YaART project page}.}.

\subsubsection{Metric.} 
Our primary evaluation metric is the Side-by-side (SbS) comparisons between two images generated from the same prompt. To compute SbS, we use a proprietary crowdsource platform with non-expert assessors. 
Before participating in the labeling process, each candidate assessor must pass an exam and achieve at least an 80\% accuracy rate among a pre-defined set of 20 assignments.
After that, we ask assessors to select one of the two images shown side-by-side based on three evaluation criteria placed in order of their importance: 

\begin{enumerate}
    \item \textbf{Defectiveness} represents the number of visual artifacts and inconsistencies on an image. Any distortions of objects, limbs, faces, and muzzles belong to this aspect.
    \item \textbf{Image-text Relevance} reflects prompt following, including the correctness of counting, sizes, and positions of objects, their relation, and properties explicitly stated in the provided textual description.
    \item \textbf{Aesthetic Quality} is responsible for image style and attractiveness. It includes color choice and balance, the beauty of the background and environment, and following the basic rules of photo composition, such as centering and the rule of thirds.
\end{enumerate}

If the images are indistinguishable based on all three criteria, they are said to have equal quality, and neither of the images gets a vote.
Each pair of images is labeled by three assessors, after which an image with more votes gets a point.
We conclude that one set of images is better than another by contesting a null hypothesis of their equality by conducting a two-sided Binomial test with a p-value of 0.05.
We do not report the FID score \cite{heusel2017_FID} because of its poor correlation with human judgments, especially for higher-quality generations.
This phenomenon was already discussed in recent papers (e.g., \cite{kirstain2024_pick_a_pic, podell2023sdxl}), and our analysis in the supplementary material confirms that.

\subsection{Scaling Experiments}
\label{sec:scaling-experiments}

In this section, we aim to answer the following questions: i) How much compute is required to train a text-to-image model to reach its quality limits? ii) Is it possible to obtain a higher-quality model by trading its size (by reducing the number of parameters) for compute time during pre-training? iii) How do model and dataset sizes affect the model training dynamic?

First, we define a set of experimental GEN64 models of different sizes: [233M, 523M, 929M, 1.45B], scaled only by width in terms of the number of channels in the convolutional layers.
All other details, such as the number of cross-attention layers, their position, etc., remain as discussed in section \ref{sec:model}.

\begin{figure}[!t]
    \centering
    \includegraphics[width=\linewidth]{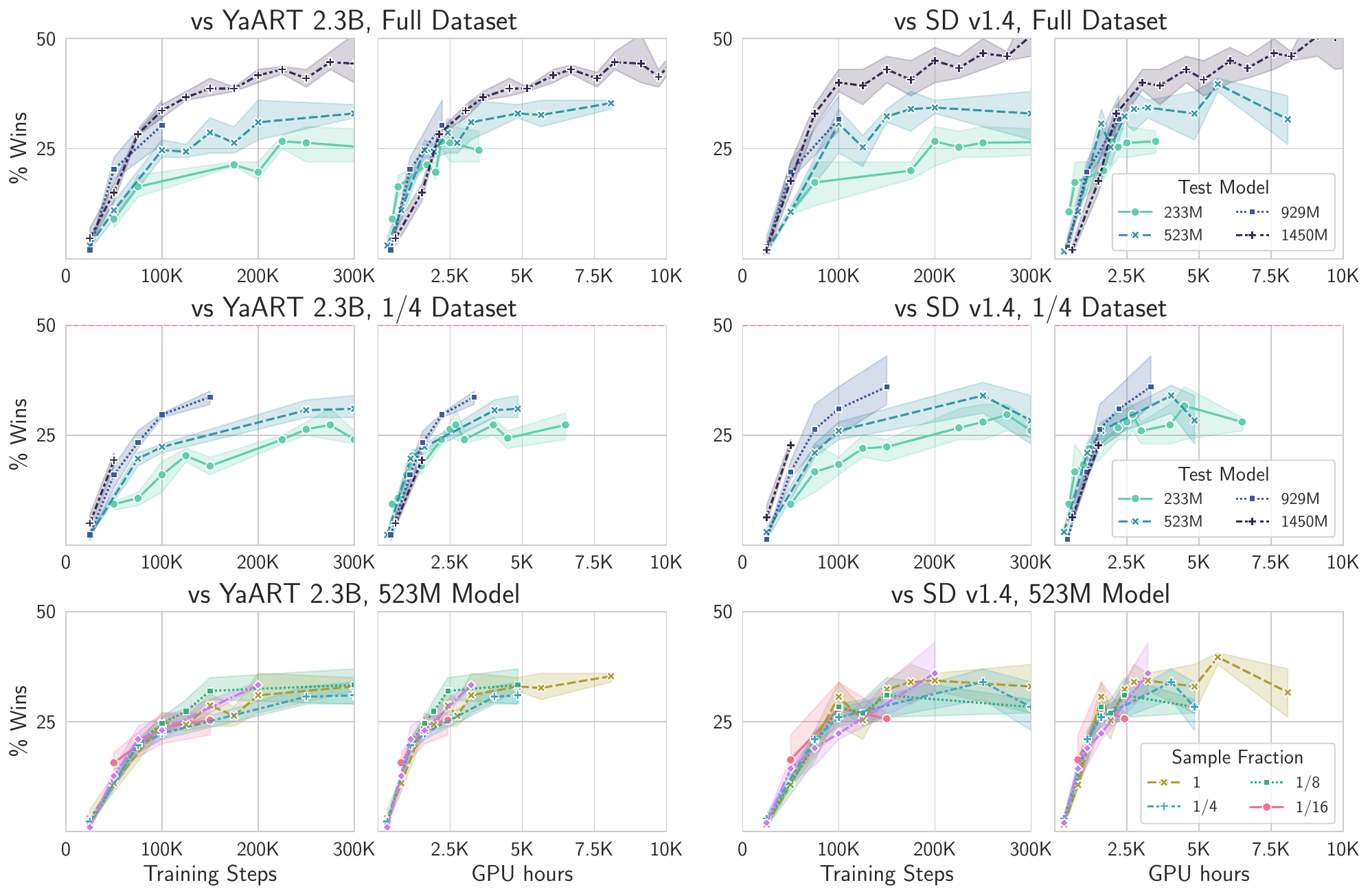}
    \caption{\textbf{Scaling up the convolutional GEN64 model improves training}, leading to higher quality models.
    Larger models train faster regarding training steps and GPU hours, leading to better results across different dataset sizes (top and middle rows).
    Dataset size weakly influences the model's end quality (bottom row).}
    \vspace{-13pt}
    \label{fig:sl_grid_main_text}
\end{figure}

\subsubsection{Learning Rate and Batch Size.}

Following the best practices from the area of Large Language Models \cite{hoffmann2022_Chinchilla}, we first aim to estimate the optimal hyperparameters (learning rate and batch size) for training models of various sizes. For that, we perform more than 100 short pre-trainings of the experimental GEN64 models with batch sizes $\in [48, 1152]$ and learning rate $\in [1 \times 10^{-8}, 5 \times 10^{-4}]$.
All runs are performed on the full 330M Main Gen dataset discussed in sec. \ref{sec:data-selection}.
By analyzing training results in terms of FID and CLIP score \cite{hessel2021clipscore} we conclude that learning rate of $1 \times 10^{-4}$ consistently leads to healthy training dynamics for larger batch sizes across all model sizes.

\subsubsection{Model Size, Dataset Size and Compute.}

\begin{figure}[!t]
    \centering
    \includegraphics[width=\linewidth]{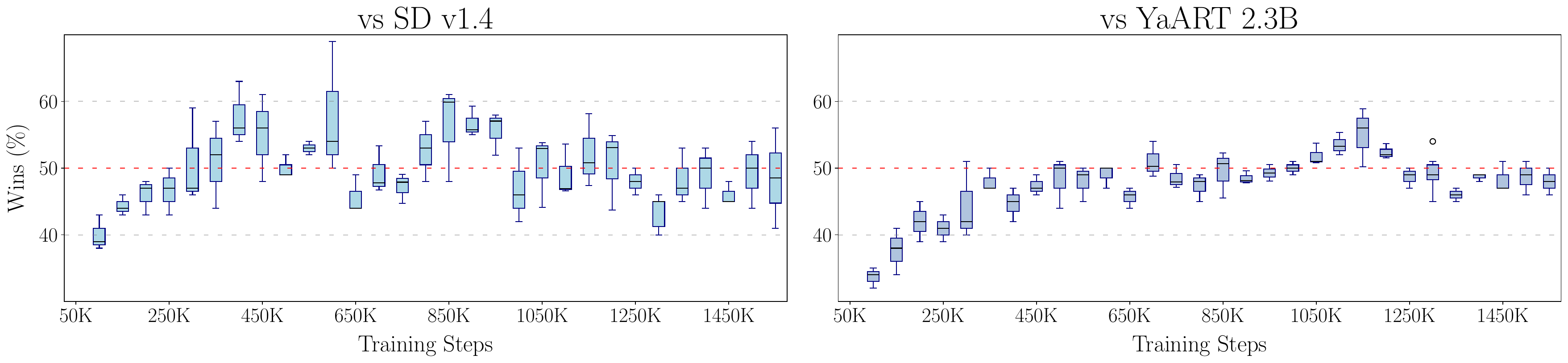}
    \caption{Dynamics of side-by-side comparisons of the half-size model with Stable Diffusion v1.4 \cite{rombach2022_LDM}  (left) and the fully sized model pre-train (right). 
    Each point shows the mean and standard deviation between three independent human evaluation experiments. 
    Note the rapid quality growth through the first few hundred iterations, after which performance reaches a plateau. 
    Given enough compute, the test model is capable of surpassing the Stable Diffusion quality, while the performance of the fully sized YaART model remains unsurpassed even with more compute.}
    \vspace{-13pt}
    \label{fig:td1_long_boxplot}
\end{figure}

Previously, DiT \cite{peebles2023_DiT} showed that scaling transformer denoiser in the LDM framework \cite{rombach2022_LDM} improves training in terms of FID. However, the question of the scalability of convolutional models in the cascade diffusion paradigm remains open. Moreover, it is unclear how dataset size is related to the model size and whether there is an optimal combination between the two.

To investigate that, we train four experimental GEN64 models defined above on five dataset scales, resulting in 20 pre-training runs.
The experimental datasets are obtained by uniform sub-sampling of the 330M Main Gen dataset with the following sampling fractions: $[1, \sfrac{1}{2}, \sfrac{1}{4}, \sfrac{1}{8}, \sfrac{1}{16}]$.
All models are trained with the same learning rate of $1 \times 10^{-4}$ and batch size of 4800, the same as for the 2.3B model pre-training.
We leave the remaining super-resolution models from the YaART cascade unchanged.

During training, each model is evaluated every 50K training steps with six SbS comparisons: three against the YaART 2.3B pre-trained model and three against SD v1.4.
We perform all evaluations on YaBasket according to the protocol described in \ref{sec:evaluation-setting}. 
Using different random seeds for image generation and random pools of human assessors allows us to regularize the evaluation procedure, avoiding potential bias towards one particular type of images.

Our experiments (Figure \ref{fig:sl_grid_main_text}) show that the quality of convolutional models grows along with the increased number of parameters. 
Also, scaling up the models makes training more efficient in terms of number of training steps and GPU hours required to achieve quality comparable to our two strong baselines: Stable Diffusion v1.4 \cite{rombach2022_LDM} and YaART 2.3B pre-training.
Remarkably, this effect persists when models are trained on datasets of different sizes.
The dataset size alone neither dramatically influences training dynamics nor leads to substantial changes in resulting model quality.
In these experiments, the models trained on the smallest train set ($\sfrac{1}{16}$ of the 330M Main Gen dataset, around 20M image-text pairs) perform about the same as models trained on the entire 330M Main Gen dataset.
This observation suggests that datasets of hundreds of millions and, sometimes, billions of image-text pairs may be excessive, and the dataset size alone is a poor predictor of the resulting model quality. 

\subsubsection{An Exchange of Model Size for Compute.}

The common wisdom in deep Learning is that scaling up a model size and computational resources typically leads to better models. 
But can we trade one for another?
To answer this question, we continue pre-training the 1.45B model from the previous section on the 330M Main Gen dataset.
Our goal is to train it for longer than the 2.3B baseline, both in terms of the number of training steps and GPU hours, to see whether it allows the test model to surpass the baseline in terms of generation quality. 

In Figure \ref{fig:td1_long_boxplot}, showing the training dynamics, we see that the visual quality relatively quickly saturates, reliably reaching the performance of SD v1.4 and YaART 2.3B pre-train around 350K and 500K iterations, respectively.
After that, the quality fluctuates for several hundred thousand iterations and even stagnates in the later stages of training. 
We conclude that scaling up the model size is an unavoidable prerequisite for further growth of generation quality.
We also see that model performance grows non-monotonically, suggesting that generation quality should be periodically evaluated throughout training.

\subsection{Data Quality vs Quantity}
\label{sec:quality-vs-quantity}

\begin{figure}[!t]
    \centering
    \includegraphics[width=.99\linewidth]{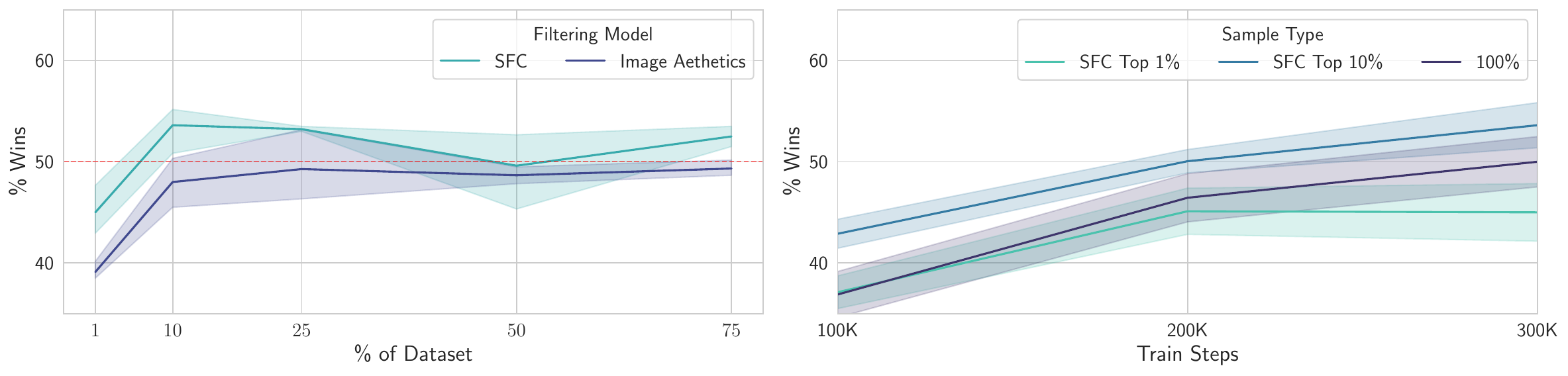}
    \caption{\textbf{Training on a small fraction of high-quality data still leads to visually appealing generations}.
    Notably, using image aesthetics as a sole criterion for data selection is constantly inferior compared to the SFC model, taking into account multiple sample quality aspects.}
    \vspace{-10pt}
    \label{fig:quality_vs_quantity}
\end{figure}

Most practical settings assume an inherent trade-off between data quality and data quantity; therefore,
the vast majority of application-oriented systems \cite{chen2023pixart, dai2023emu, podell2023sdxl, arkhipkin2023kandinsky3} perform some variation of data selection and filtering during the dataset collection.
However, the influence of the balance between data quality and quantity on the text-to-image model training dynamics remains unexplored.
In this section, we aim to investigate how a gradual decrease in dataset size in favor of selecting higher-quality samples affects the dynamics of model pre-training.
We base our analysis on a comparison of the two approaches for sample ranking used for the initial data filtering: the SFC model based on 56 image-text factors and the Image Aesthetics model.

Starting from the 330M Main Gen dataset, we progressively drop a fraction of the worst samples according to one of the aforementioned models. We then train 2.3B models for 300K iterations with default hyperparameters.
We use SbS comparisons to compare the resulting models with the baseline trained on the full dataset for the same 300K iterations.

Figure \ref{fig:quality_vs_quantity} shows that the sub-sampling dataset with the Image Aesthetics classifier only monotonically reduces the quality of the final model, matching with the intuition that larger datasets might lead to better-performing models.
It also suggests that removing image-text pairs only by means of image aesthetic attractiveness is suboptimal.

At the same time, sub-sampling data according to the SFC model is more nuanced.
Our evaluations suggest that the best results are obtained at 10\% dataset size, 33M samples.
Further reduction of the dataset size employing more strict filtering leads to a drop in generation quality.
There is also a limit on the minimum size of the dataset, as we show that scaling down to 1\% significantly hurts the performance.
%

\subsection{Effect of the model performance on fine-tuning quality.}

\begin{figure}[!t]
  \centering
  \includegraphics[width=0.9\textwidth]{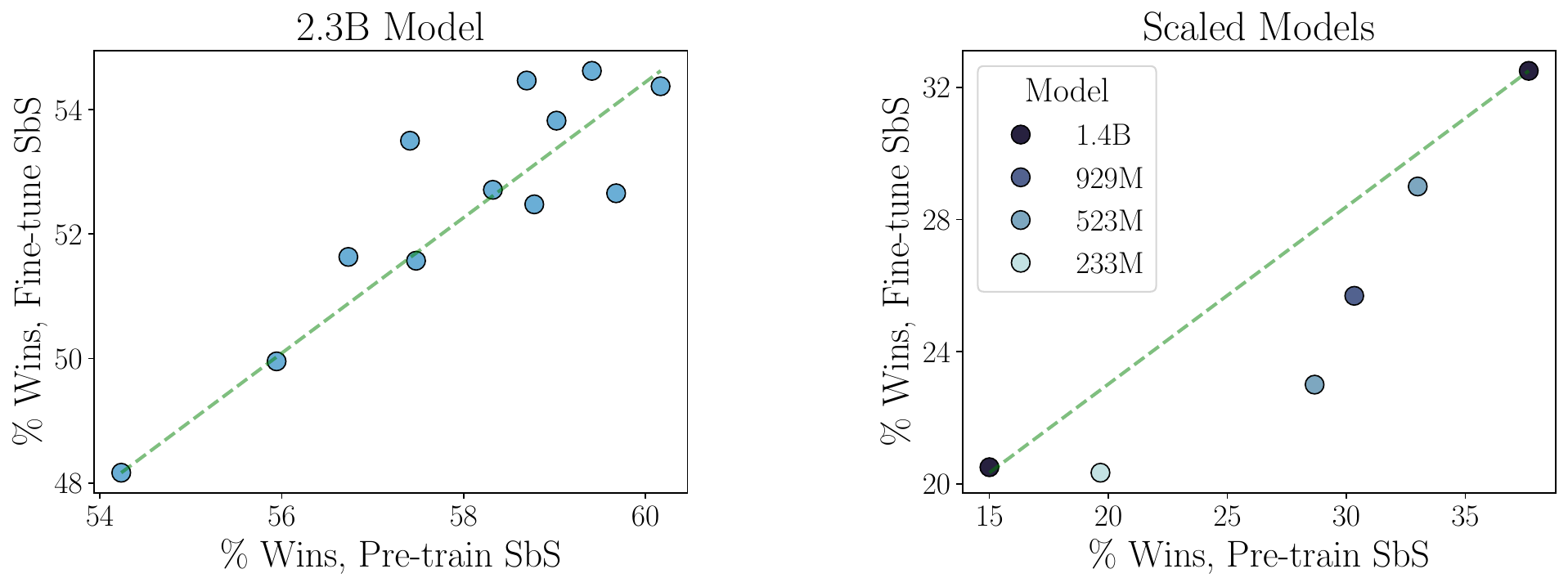} 
  \caption{\textbf{Pre-train quality strongly correlates with fine-tune quality} both for well-trained models performing similarly with YaART 2.3B pre-train (left) and smaller models from the Scaling Experiments (right).}
  \label{fig:pretrain-finetune}
  \vspace{-13pt}
\end{figure}


In previous experiments, we aimed to understand the influence of various factors on training dynamics and the quality of the model after the pre-training stage. However, from the practical perspective, one typically aims to maximize the final quality, i.e., after fine-tuning the pre-trained model. However, the relation between quality after these two stages of training is unclear. To investigate that, we fine-tune various quality models from the Scaling Experiments \ref{sec:scaling-experiments} following the procedure described in Section \ref{sec:fine-tuning}.

Figure \ref{fig:pretrain-finetune} summarizes the results of SbS comparisons between pre-trained models and their fine-tuned versions.
We observe a strong correlation between performances in these two stages of training despite variations in model size, dataset size, and dataset quality. 

%% file: sections/10_conclusion.tex
\section{Conclusion}
\label{sec:conclusion}

In this paper, we presented YaART --- a production-grade diffusion model for text-to-image generation. We show that the choices of model and dataset sizes, as well as the quality of the training images, are important and interconnected degrees of freedom that should be accurately specified for the optimal exploitation of the available GPU power. Also, we describe the procedure of tuning via Reinforcement Learning for YaART, thus confirming that RL is beneficial for production-level diffusion models.

%% file: supmat/01_intro.tex
\newpage
\section*{Supplementary Material}
\vspace{-3pt}

\begin{figure}[b]
  \centering
  \includegraphics[width=0.9\linewidth]{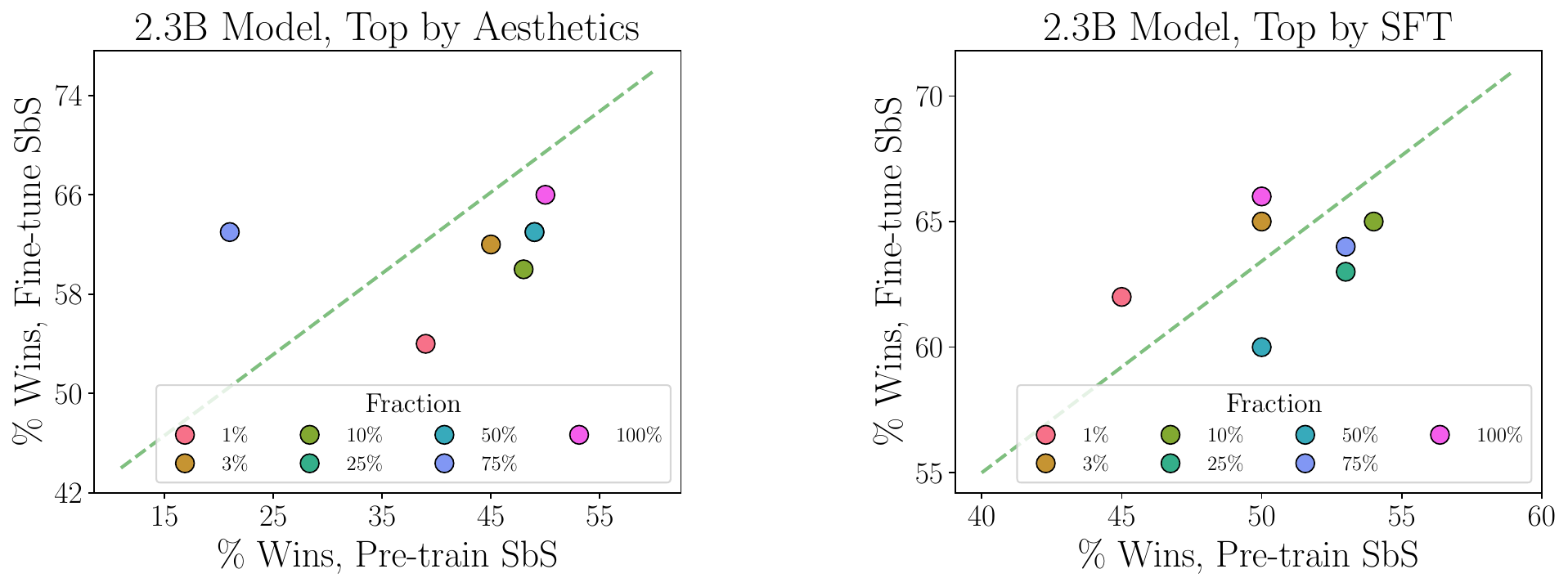} 
  \caption{\textbf{Pre-train quality strongly correlates with fine-tune quality when trained on a fraction of higher-quality data}.
  This effect is consistent across two data quality estimation model types: Image Aesthetics (left) and SFT (right).}
  \label{fig:pretrain_finetune_qvq}
\end{figure}

Here we show additional results on the correlation between pre-train and fine-tune performance for quality vs quantity experiments (Figure  \ref{fig:pretrain_finetune_qvq}).
We make additional comparisons with state-of-the-art on DrawBench (Table \ref{table:sbs-baselines-drawbench}) and provide a detailed description of hyperparameters (Table \ref{tab:model_details}).
Finally, we discuss YaART's limitations and provide additional examples of generated images, showcasing its high visual consistency, aesthetic attractiveness, and strict prompt-following abilities.

%% file: supmat/05_pretrain_ft_qvq.tex
\subsection*{Effect of the model performance on fine-tuning for Quality vs Quantity trade-off}
\vspace{-3pt}
\label{sup:pretrain_ft_qvq}

We previously discussed that pre-train quality strongly correlates with fine-tune quality in terms of our three-aspect side-by-side (SbS) comparisons regardless of the model size given a full 330M Main Gen dataset for training.
This experiment aims to understand whether the dataset size and quality introduce any difference.
For that, we fine-tune checkpoints from the Quality vs Quality experiment, i.e., 2.3B GEN64 models trained on samples of the best image-text pairs according to Image Aesthetics and Sample Fidelity Classifier (SFC) models.

Figure \ref{fig:pretrain_finetune_qvq} shows that even models pre-trained on small sample fractions (1\%, 3\%) are easily exposed to fine-tuning.
However, even though the best pre-trained model was obtained by training on the Top 10\% of data according to the SFT model, the model trained on the entire dataset still achieves the best fine-tuning quality. 
This suggests that having a large pre-train set of moderate-quality data is a strong prerequisite for achieving high-end quality.

\input{tables/sbs_baselines_drawbench}

%% file: tables/sbs_baselines_drawbench.tex
\begin{table}[t]
\setlength\tabcolsep{2.5pt}
\begin{center}
\resizebox{.9\columnwidth}{!}{ 
\begin{tabular}{l@{\hskip .3in}c@{\hskip .2in}c@{\hskip .2in}c@{\hskip .4in}c}
\toprule

&\multicolumn{4}{c}{Quality Aspects}  \\
 \cmidrule{2-5} 
 YaART 2.3B RL & Beauty & Defects & Alignment &  Overall \\
  \cmidrule{1-5}
  vs MidJourney v5 & $0.58\pm0.01$ & $0.54\pm0.01$ & $0.52\pm0.02$ & $\textbf{0.55}\pm \textbf{0.03}$  \\
  vs SDXL & $0.78\pm0.04$ & $0.76\pm0.03$ & $0.54\pm0.02$ & $\textbf{0.82}\pm \textbf{0.02}$ \\
  vs Kandinsky v3 & $0.61\pm0.03$ & $0.70\pm0.02$ & $0.48\pm0.02$ & $\textbf{0.73}\pm\textbf{0.02}$ \\
  vs OpenJourney & $0.86\pm0.06$ & $0.80\pm0.01$ & $0.82\pm0.05$ & $\textbf{0.94}\pm\textbf{0.01}$ \\
  
\bottomrule
\end{tabular}
}
\end{center}

    
\caption{We compare YaART with state-of-the-art based on three main evaluation criteria and overall human preferences on DrawBench.
\textbf{Bold} denotes a statistically significant overall model quality difference.}

\label{table:sbs-baselines-drawbench}
\end{table}

%% file: supmat/06_limitations.tex
\section*{Limitations}
\label{sup:limitations}

\input{tables/model_details}

Contrary to the mainstream approach to training text-to-image diffusion models as LDMs, we have chosen the cascaded diffusion variant. 
Although this approach demands more computational resources at runtime, its interactive nature presents significant benefits: we can provide a 256x256 image that users can further filter and refine.
Moreover, we apply the super-resolution stage only to a selected fraction of these images.
Despite all the recent progress in this area, modern diffusion models still require substantial human supervision. This typically involves iterative prompt modification, sampling parameter tuning (or even adjusting the random seed), and a post-filtering process for the generated images.

We have also applied automatic filters to eliminate images containing text from our dataset. 
This decision stems from our belief that the quality of text generation currently falls short of practical standards.
Interestingly, despite these limitations, our model has successfully learned to generate reasonable characters and words from the few text-containing images in our training dataset.


    

%% file: tables/model_details.tex
\begin{table}[t]
\centering

\scalebox{.9}{
\begin{tabular}{l@{\hskip .2in}c@{\hskip .2in}c@{\hskip .2in}c@{\hskip .2in}c@{\hskip .2in}c}
\toprule
 & 64 & 64 $\rightarrow$ 256 & 256 $\rightarrow$ 1024 \\
\midrule
Noise Schedule & \emph{linear} & \emph{linear} & \emph{linear} \\
Sampling Steps & 32 & 32 & 32 \\
Crop Fraction & 1 & 1 & \sfrac{1}{4} \\
Model Size & 2.3B & 700M & 700M \\
Channels & 448 & 320 & 128 \\
Depth & 4 & 4 & 5  \\
Channels Multiplier & $[1, 2, 3, 4]$ & $[1, 2, 3, 4]$ & $[1, 2, 2, 4, 8]$  \\
Heads Channels & 64 & 64 & - \\
Attention Resolution & $[2, 4, 8]$ + middle & middle & -  \\
Text Encoder Context & 128 & 128 & uncond  \\
Text Encoder Width & 1536 & 1536 & uncond  \\
Dropout & 0 & 0 & 0 \\
Weight Decay & 0 & 0 & 0 \\
Batch Size & 4800 & 960 & 512 \\
Iterations & $1.1 \times 10^6$ & $1.5 \times 10^6$ & $1.5 \times 10^6$ \\
Learning rate & $1 \times 10^{-4}$ & $5 \times 10^{-5}$ & $5 \times 10^{-5}$ \\
Adam $\beta_1$ & 0.9 & 0.9 & 0.9 \\
Adam $\beta_2$ & 0.99 & 0.99 & 0.99 \\
EMA Decay & 0.9999 & 0.9999 & 0.9999 \\
\bottomrule
\end{tabular}
}
\bigskip
\caption{
Hyperparameters for the models in the YaART cascade.
}
\vspace{-23pt}
\label{tab:model_details}
\end{table}

%% file: supmat/07_more_examples.tex

\begin{figure}[tp]
  \centering
  \makebox[\textwidth][c]{\includegraphics[width=1.5\linewidth]{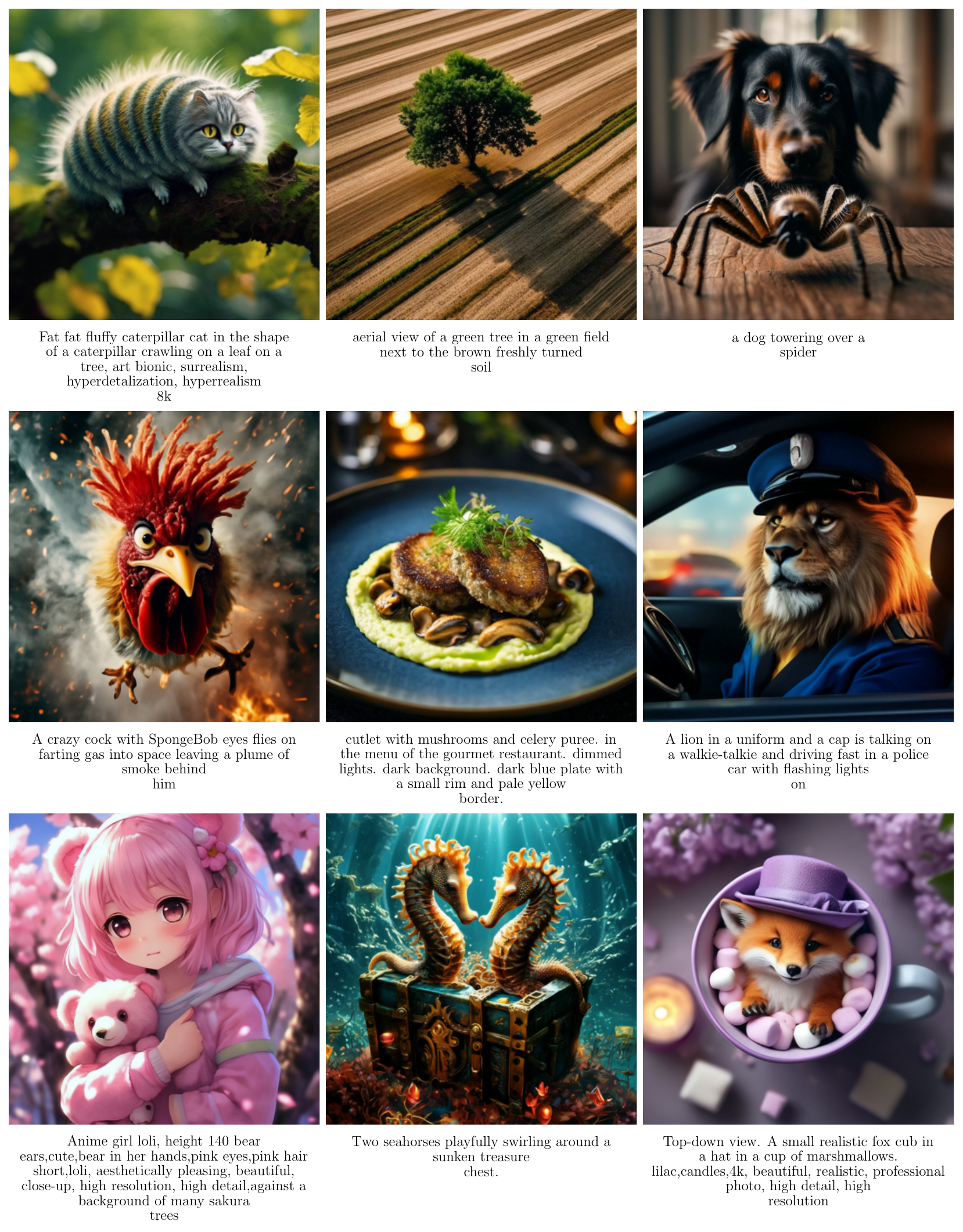} }
\end{figure}

\begin{figure}[tp]
  \centering
  \makebox[\textwidth][c]{\includegraphics[width=1.5\linewidth]{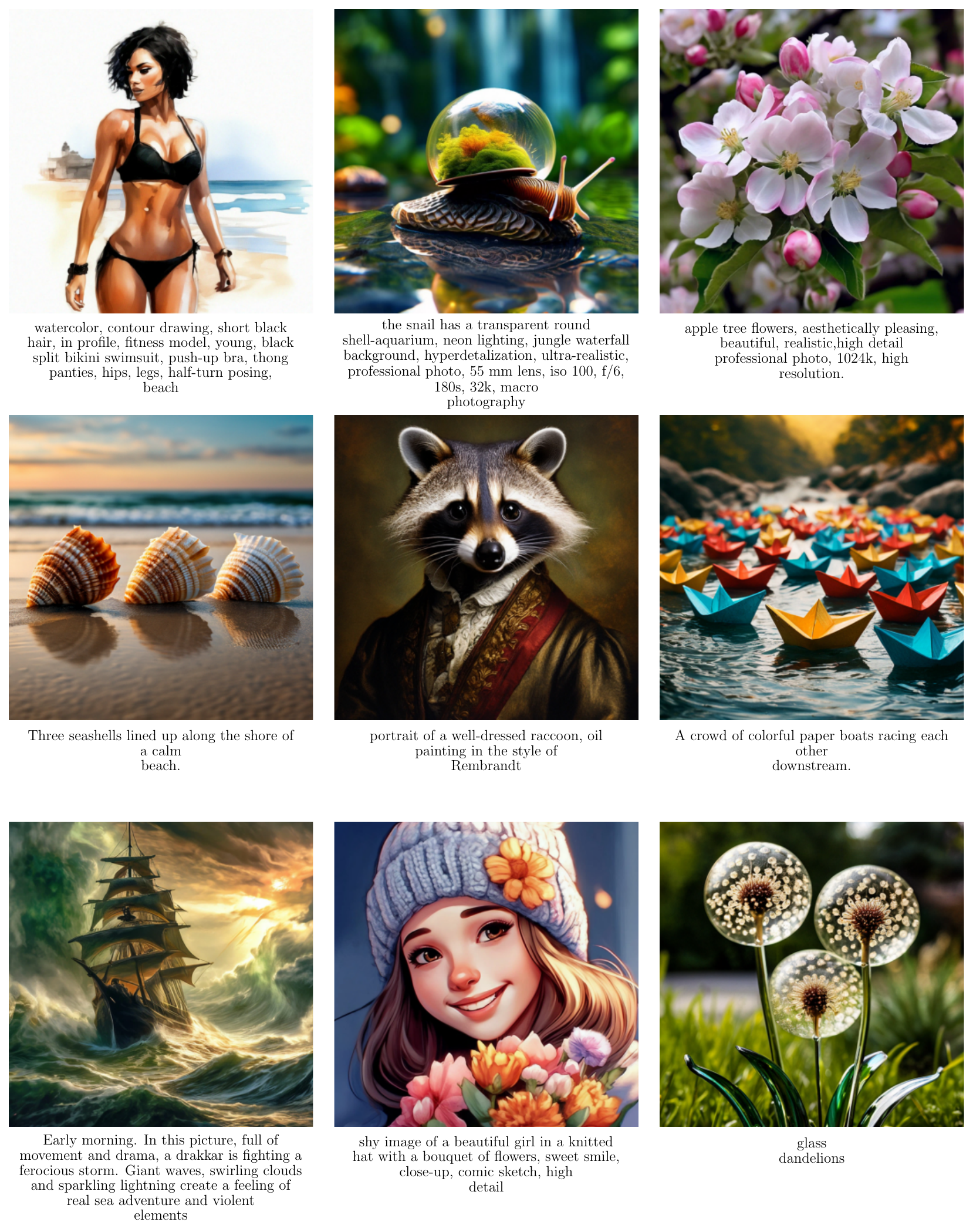}} 
\end{figure}

\begin{figure}[tp]
  \centering
  \makebox[\textwidth][c]{\includegraphics[width=1.5\linewidth]{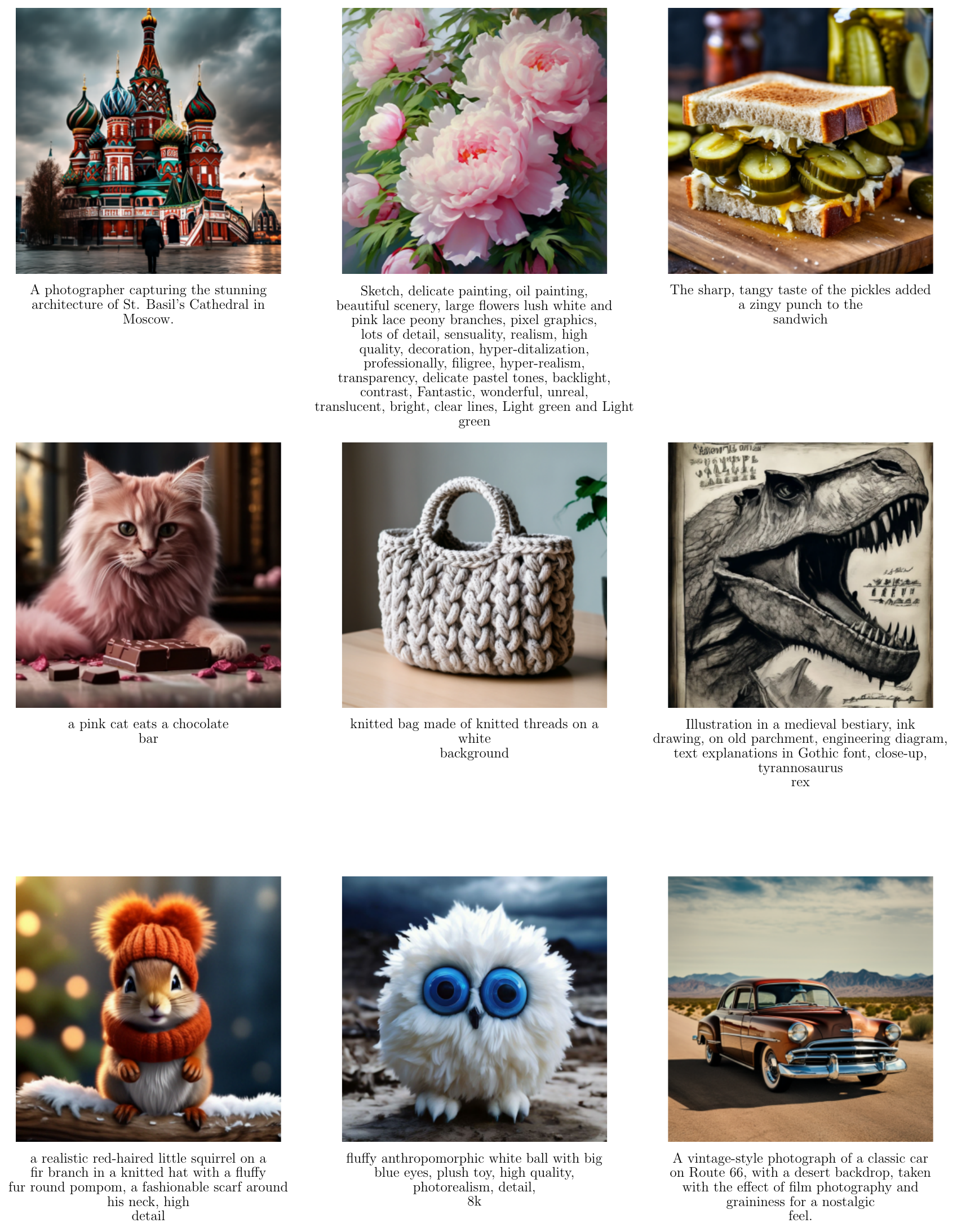}} 
\end{figure}

\begin{figure}[tp]
  \centering
  \makebox[\textwidth][c]{\includegraphics[width=1.5\linewidth]{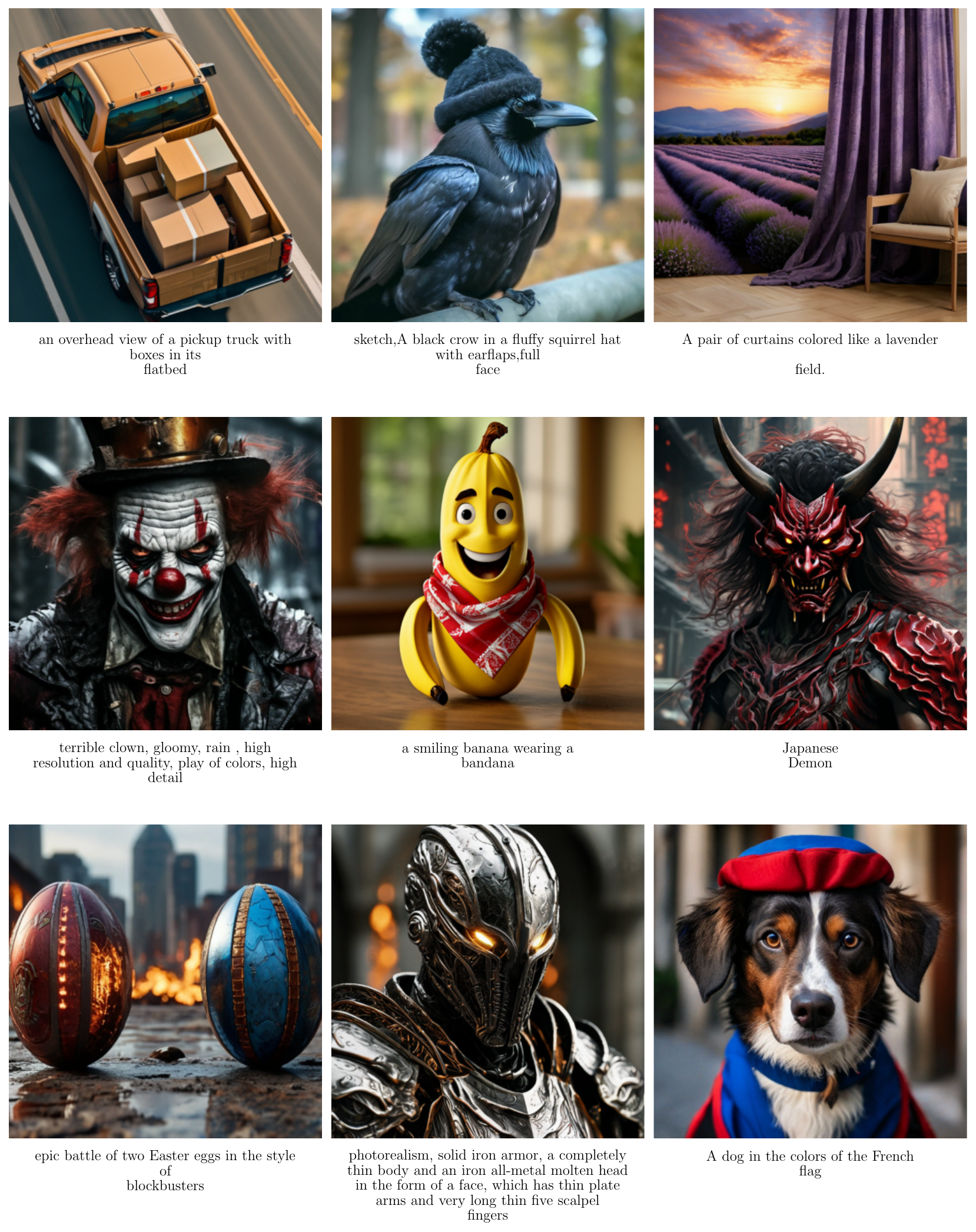}} 
\end{figure}